\documentclass{cta-author}
\usepackage{hyperref}
\usepackage{natbib}
\usepackage{tabularx}
\usepackage{booktabs}
\usepackage{makecell}
\usepackage{threeparttable}
\usepackage{url}

{}
{}
{}

\begin{document}

\title{Short-Term Electricity Price Forecasting based on Graph Convolution Network and Attention Mechanism}

\author{\au{Yuyun~Yang$^{1}$},  \au{Zhenfei~Tan$^{1}$}, \au{Haitao~Yang$^{1}$}, \au{Guangchun~Ruan$^{1}$}, \au{Haiwang~Zhong$^{1}$\corr}}

\address{\add{1}{State Key Laboratory of Power Systems, Department of Electrical Engineering, Tsinghua University, Beijing 100084, China}
\email{zhonghw@mail.tsinghua.edu.cn}}

\begin{abstract}
In electricity markets, locational marginal price (LMP) forecasting is particularly important for market participants in making reasonable bidding strategies, managing potential trading risks, and supporting efficient system planning and operation. Unlike existing methods that only consider LMPs' temporal features, this paper tailors a spectral graph convolutional network (GCN) to greatly improve the accuracy of short-term LMP forecasting. A three-branch network structure is then designed to match the structure of LMPs' compositions. Such kind of network can extract the spatial-temporal features of LMPs, and provide fast and high-quality predictions for all nodes simultaneously. The attention mechanism is also implemented to assign varying importance weights between different nodes and time slots. Case studies based on the IEEE-118 test system and real-world data from the PJM validate that the proposed model outperforms existing forecasting models in accuracy, and maintains a robust performance by avoiding extreme errors.
\end{abstract}

\maketitle

\section{Introduction}
\subsection{Background}
With deregulation in global power industries over the last several decades, more than 30 countries/territories have established their electricity markets~\cite{griffin2009electricity}, while many developing countries also initiated electricity market reforms recently~\cite{tan2018security}. In most of these markets, the locational marginal price (LMP) mechanism is widely applied owing to its certified features of incentive compatibility, revenue adequacy, cost causation-awareness, and transparency \cite{lmp:feature}. Therefore, LMP forecasting helps market participants and system operators in various decision-making scenarios, including 
multi-area market coordination~\cite{wang2019incentive}, 
energy sharing~\cite{chen2021communication}, 
bidding~\cite{ruan2020constructing}, 
microgrid dispatch~\cite{zhou2020forming}, 
energy storage scheduling~\cite{fang2016strategic},
and building energy management~\cite{zhang2020soft}.

Among all prediction tasks in power systems, short-term LMP forecasting is more difficult than its competitors, e.g., load forecasting, renewable generation forecasting, and the reasons mainly lie in three aspects.
First, LMPs are influenced by much more factors and thus become more volatile~\cite{veeramsetty2020probabilistic}. A lower prediction accuracy could often be expected when forecasting LMPs.
Second, LMPs rely heavily on commercial bidding strategies of market participants~\cite{saebi2010demand}, which are private information and hard to collect.
Third, LMPs at different locations are often spatially related due to the system network connection~\cite{litvinov2004marginal}, so time-series analysis alone may not be enough. 

We pay special focus to the third reason and point out that ignoring LMPs' spatial inter-dependency could restrict the forecasting accuracy. 

Most existing methods treat LMPs at different nodes as independent time series, and it remains challenging on how to integrate the spatial correlation in LMP forecasting.

\subsection{Related works}
Time-series models are applicable for LMP forecasting and often exhibit understandable intuitions and physical interpretations. 
Reference~\cite{koopman2007periodic} formulated some general seasonal periodic regression models for daily LMPs with the auto-regressive integrated moving average (ARIMA), auto-regressive fractionally integrated moving average, and generalized auto-regressive conditional heteroskedasticity (GARCH) disturbances. 
Reference~\cite{cuaresma2004forecasting} applied variants of the auto-regressive (AR) model and general auto-regressive moving average (ARMA) processes (including ARMA with jumps) to predict short-term electricity prices in Germany. 
An AR model with exogenous variables was implemented for day-ahead LMP forecasting in \cite{chitsaz2017electricity}, while \cite{pircalabu2017regime} put forward a regime-switching AR–GARCH copula to discover the joint behavior of day-ahead electricity prices in interconnected European markets. 
Reference~\cite{ruan2020neural} implemented a seasonal ARIMA model to generate the LMP scenario set which was informative to show the price uncertainty.
The extended ARIMA approach in~\cite{zhou2006electricity} was able to estimate the confidence intervals of the predicted LMPs.
In~\cite{contreras2003arima}, the day-ahead LMP forecasting could reach an average 11\% weekly mean absolute percentage error (MAPE).

With the booming of various machine learning techniques~\cite{ruan2020review}, more and more researchers have shifted their focus to this new direction. In general, the machine learning models are advantageous to handle complex and nonlinear correlations~\cite{weron2014electricity}, making it become a powerful toolbox for tough forecasting tasks. 
A support vector machine model was developed to predict LMPs in~\cite{wu2006forecasting}. 
Then \cite{fan2007next} established a two-stage hybrid network of self-organized map (SOM) and support vector machine (SVM).
Reference~\cite{mandal2012hybrid} provided a single-node LMP forecasting model based on the artificial neural network. 
Reference~\cite{mandal2007novel} implemented a multi-layer perceptron model and finally achieved a 7.66\% day-ahead MAPE. 
An enhanced probability neural network was utilized in day-ahead LMP forecasting, whose daily MAPE was 5.36\% \cite{lin2010electricity}. 
Similar models include the feed-forward nonlinear MLP~\cite{aggarwal2009electricity,rodriguez2004energy} and the recurrent neural network (RNN)~\cite{zhang2020deep}. 
Here, most neural network models considered the hourly predictions of single-node LMPs, and the typical daily MAPE reached around 6\%~\cite{lee2005system}. 

In recent years, deep learning models are verified to greatly improve forecasting performance.
Reference~\cite{luo2019two} applied the deep neural network (DNN) and support vector regression to predict real-time market prices, and its mean square error reached 20.51~${\$/MWh}^2$. 
The radial basis function networks~\cite{lin2010enhanced} (better performance on price spikes) and recurrent neural networks~\cite{mandal2010new} could achieve day-ahead MAPEs of 5.56\% and 7.66\% respectively. 
Reference~\cite{chang2019electricity} compared the performance of traditional AR models with deep learning approaches, e.g., long-short term memory (LSTM) networks. 
Reference~\cite{lago2018forecasting} took a step forward by formulating a hybrid DNN-LSTM network to simultaneously predict day-ahead prices in several countries. This network achieved 13.06\% of symmetric mean absolute percentage error (sMAPE). 
Reference~\cite{zheng2020locational} stacked the decision tree regressor, random forest regressor, and extremely randomized tree regressor to predict different components of LMPs, and achieved the best mean absolute error (MAE) of 4.08~$\$/MWh$.
In~\cite{afrasiabi2019probabilistic}, the price forecast procedure consisted of a convolution neural network (CNN), gated recurrent unit (GRU), and adaptive kernel density estimator. 

In the above references, LMPs at different nodes are predicted independently, failing to extract the inherent spatial correlations among different locations, especially in congestion situations. It is generally believed that the spatial distribution of LMPs follows certain modes, but very limited attention is paid to this aspect, and simultaneous prediction of LMPs at multiple locations has not received much attention.

The recent graph convolutional network (GCN) offers a way to incorporate the spatial correlation of LMPs in a neural network. 
The GCN model was first proposed in \cite{kipf2016semi} to broaden the traditional convolution networks to incorporate graph-structured data. GCN is thus a promising option candidate to capture the topological relationship of LMPs at different locations.
Currently, the GCN was successfully applied in traffic flow forecasting~\cite{guo2019attention}. 
GCNs were also utilized to forecast wind power generation incorporating spatial correlation in~\cite{wang2008security}, and solve unit commitment and economic dispatch problems~\cite{gaikwad2020using}. However, there was no paper that considered LMP forecasting with GCNs so far. 

It should also be pointed out that classical GCNs cannot capture the temporal correlations, which is a simple task for various recurrent neural networks, e.g.,~\cite{zhang2020deep,lago2018forecasting}. However, simultaneous consideration of temporal and spatial correlation of LMPs has not been reported in the existing works of literature.

To this end, we propose a novel model based on the GCN and temporal convolution (namely ST-Conv), and a useful technique, attention mechanism, is also employed. The major contributions of this paper are summarized as follows:
\begin{itemize}
    \item Different from the existing works, the proposed model can capture the spatial and temporal features of LMPs simultaneously. The spatial correlation is encoded with a spectral graph convolutional network by modeling the electric grid as an undirected graph, while the temporal correlation is captured by a one-dimensional convolutional network.
    \item The proposed model can handle all LMPs of different locations simultaneously, and the spatial correlation is thus accurately and efficiently stored in this pattern.
    \item The attention mechanism is implemented to guide the model to distinguish and focus more on the important input information.
\end{itemize}

The remainder of this paper is organized as follows. Section~\ref{framework} introduces the overall forecasting framework, Section~\ref{model} elaborates the key techniques, including the GCN implementation, temporal convolution, and attention mechanism. More details about how the forecasting model is trained are clarified in Section~\ref{trainingmethod}. Several simulation results are discussed in Section~\ref{case}. At last, Section~\ref{conclusion} draws the conclusions.

\section{Framework}
\label{framework}
\subsection{LMP derivation and compositions}
\label{GCN_LMP}

The LMP for a network node is defined as the incremental operating cost to supply another megawatt of power at this node. Mathematically, the LMP is the optimal dual variable of the economic dispatch problem. The economic dispatch problem optimizes the total welfare of the whole system, which equals the total utility of consumers minus the total generation cost of generators. The LMP at each node consists of three parts: the energy component $\lambda \in \mathbb{R}$, congestion component $\mu \in \mathbb{R}^N$, and network loss component $\nu \in \mathbb{R}^N$ \cite{litvinov2004marginal}. The network loss accounts for a minor percentage of LMP, so we ignore this component in the prediction. Then LMP can be represented as the sum of energy component $\lambda$ and congestion component $\mu$. When the congestion occurs, LMPs at different nodes will deviate from each other since congestion components become non-zero. Even though, the spatial distribution of the LMP follows certain modes. The congestion component of the LMP takes the form of a linear combination of power transmission distribution factors (PTDF) of congested transmission lines and thus, LMPs at different nodes are restricted to an affine subspace, which implies the necessity to put emphasis on the spatial correlation of LMPs.

Take LMP at node $i$ for instance,
\begin{align} 
\label{equ:lmp}
L M P_{i} = \lambda+\mu_i = \lambda+\sum_{k=1}^{m} t_{W: i, k} \mu_{k}, 
\end{align} 
where $t_{w: i, k}$is the constraint $k$’s power flow sensitivity to the injection at node $i$ concerning the slack reference $W$, and $m$ is the number of constraints.

From the view of power transfer distribution factor (PTDF), a certain congestion component $\mu$ represents a dual multiplier by the restrictions of its corresponding power line, which implies that the congestion components are restricted to the row space of PTDF. That means LMP would be affected by the topological connections between nodes. 

Traditional neural-network-based models directly give LMP predictions node by node according to its historical information. However, in this work, we try to forecast LMP at all nodes simultaneously. This leads to an additional challenge that models can hardly give strict zero-$\mu$ at all nodes when there is no congestion. Thus, we provide a novel LMP decomposition: $LMP=\lambda+s\mu$, where congestion factor $s \in \{0,1\}$. The model additionally proposes a binary prediction of $s$. When $s=0$, there is no congestion in the system, so LMP at all nodes equals $\lambda$; otherwise when $s=1$, indicating congestion occurs and all non-zero dual multipliers should contribute to $LMP$, and $LMP = \lambda + \mu$. In this way, we are able to give precise predictions at all nodes within one procedure.

Unlike existing methods that take the temporal features of LMP into account only, the proposed method based on Graph Convolutional Network (GCN) considers a power system as an undirected graph and builds a spectral graph convolutional network to extract the spatial features of LMP. Furthermore, we propose an attention mechanism and temporal convolution to enhance its performance. The GCN-based method can forecast the LMP of all nodes simultaneously, making up for the defect that traditional neural networks can not accurately restore the spatial information of LMP.

\subsection{Structure of the LMP forecasting model}
\label{structure}
\begin{figure}[htbp] 
  \centering
  \includegraphics[width=1.0\linewidth]{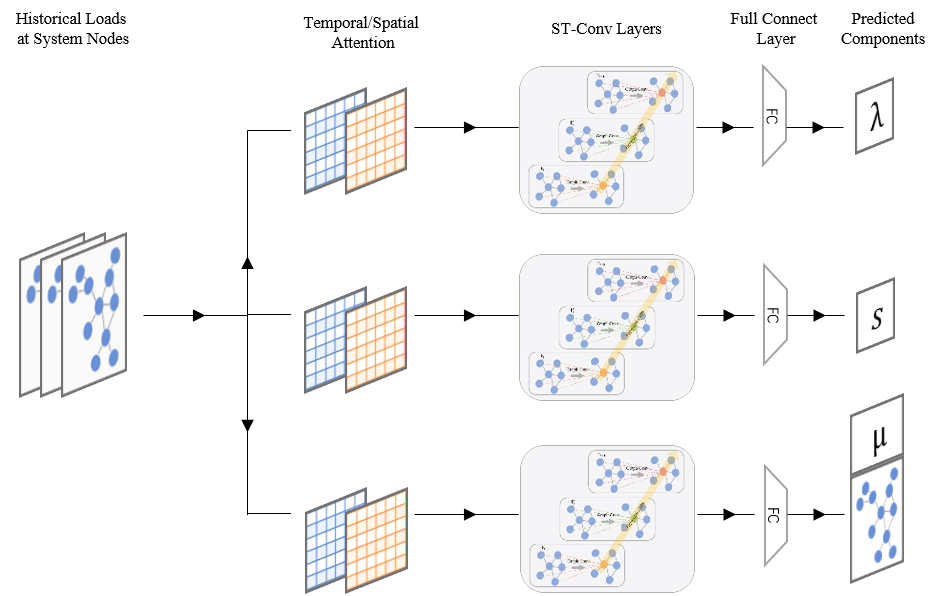}
  \caption{Framework of LMP-Forecasting ASTGCN}
  \label{fig_astgcnbranch}
\end{figure}

The proposed forecasting networks (GCN and its descendant: Attention-based Spatial-Temporal Graph Convolutional Network, ASTGCN) share the same structure of three similar branches (Fig.\ref{fig_astgcnbranch}), respectively forecasting different components $\lambda, s,~and~\mu$. For each branch, the input is always the historical loads of nodes in a topological structure. Then it goes through a temporal/spatial attention layer and the loads' value will be re-weighted according to their importance for prediction. After that, two identical consecutive layers of graph-temporal convolution would convolve nodes and their neighbours to get internal information. Finally, the full connect layer formulates the required shape of $\lambda, s,~and~\mu$ as output. Their composition  $LMP = \lambda + s\mu$ produces the predicted LMP. In different branches, we set corresponding parameters as Tab.\ref{tab:ASTGCN_params} shows. The value of $q$ depends on which branch the layer is in. For $\lambda,~q=1$, indicating an only value is determined for the energy component; for $s,~q=2$, indicating the possibility of congestion or not; for $\mu,~q=16$, using a 16-dimension tensor to represent characteristics of congestion. The detailed structure of ASTGCN will be introduced in the following sections. 

\begin{table}[htbp]
	\centering
	\begin{threeparttable}[b]
	\caption{Parameters of layer shape}
	\label{tab:ASTGCN_params}
	\begin{tabular}{ccccc}
		\toprule[1.5pt]
		& input dimension & output dimension & K\tnote{1} & T\tnote{2}\\
		\midrule[1pt]
		Attention layer  & $T\times N$ & $T\times N$ & $/$ & $/$\\
		ST-Conv 1  & $T\times N$ & $T\times N\times128$ & $2$ & $3\times 1$ \\
		ST-Conv 2  & $T\times N\times128$ & $T\times N\times128$ & $2$ & $3\times 1$\\
		Full connect  & $T\times N\times128$ & $1\times N\times q $ & $/$ & $/$\\
		\bottomrule[1.5pt]
	\end{tabular}
    \begin{tablenotes}
     \item[1] Graph receptive field
     \item[2] Temporal kernel's shape
   \end{tablenotes}
    \end{threeparttable}
\end{table}

\section{Model}
\label{model}

The proposed forecasting model aims to solve this problem: Given $\chi=\left(X_1,X_2,\cdots,X_n\right)\in\mathbb{R^{N}}$, which denotes historical power loads of all the nodes in the power system, it needs to forecast LMP at all nodes for the same moment. We assumed that the power system contains $N$ nodes, and for each prediction, load data of previous $T$ Hours of loads are available.

\begin{figure}[htbp]
  \centering
  \includegraphics[width=1.\linewidth]{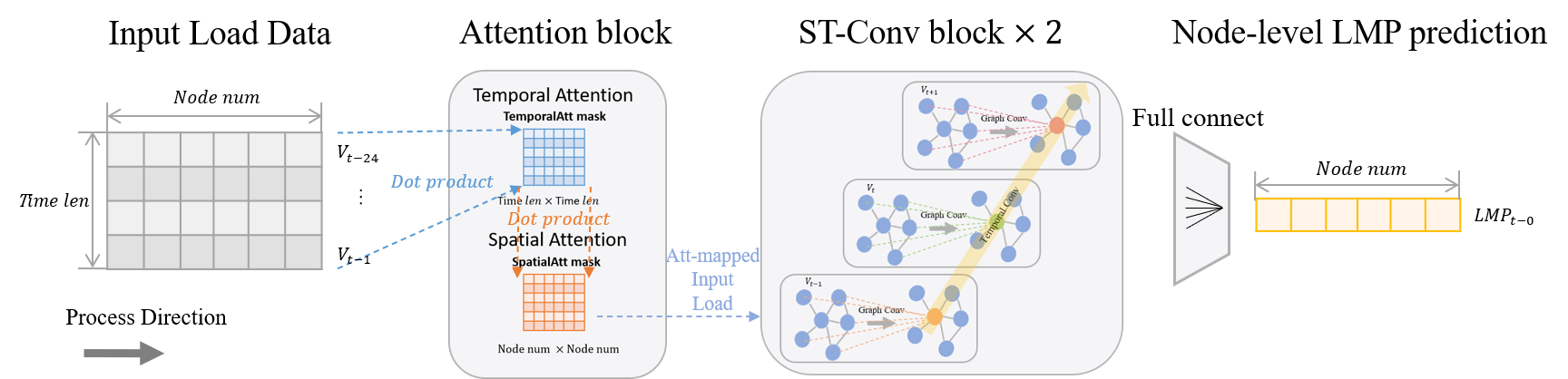}
  \caption{One branch of LMP-Forecasting ASTGCN}
  \label{fig_zoomast}
\end{figure}

A detailed zoom-in of one forecasting branch for instance is shown in Fig.\ref{fig_zoomast}. Input load data comprising history loads at all nodes first come through a pre-trained attention block and thus become a mapped input. The attention block contains two masks respectively designed for time and space pattern extraction. After that, the mapped input will be processed by two consecutive ST-Conv blocks (made up of Graph and Temporal Convolution). Finally, a Fully Connect layer would synthesize the patterns learnt from the input and give an LMP prediction of all nodes for the next future moment.

\subsection{Input}
The initial input of the whole model contains the historical loads(MWh) at all nodes structured in table format. In the studied case, the previous hourly loads of $T$ hours are used as an input, whose shape is of $T~hr \times N~nodes$.

\subsection{Attention Block}
\label{attmech}
\begin{figure}[htbp]
  \centering
  \includegraphics[width=1\linewidth]{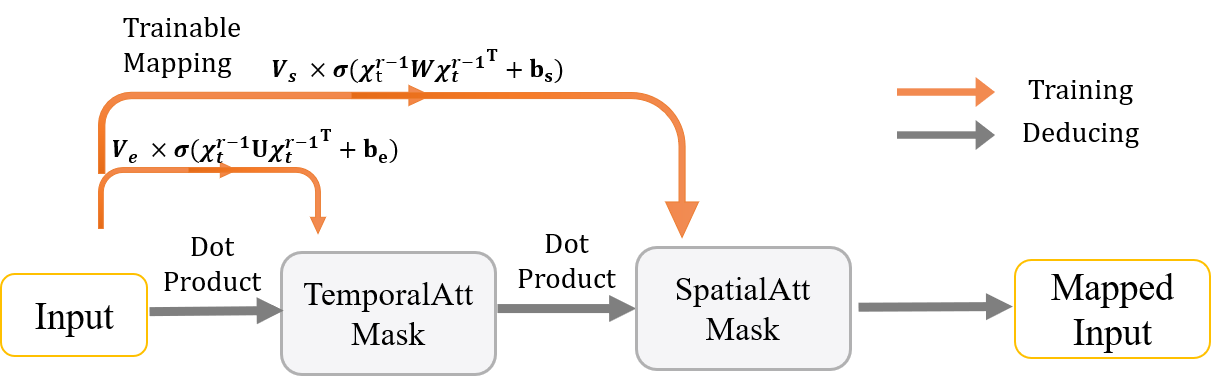}
  \caption{Training and Deducing of Attention Mapping}
  \label{fig:attProc}
\end{figure}

The attention block consists of two masks corresponding to time and space dimension, namely spatial attention and temporal attention, to help adaptively identify the correlations between different time-points/system-nodes to reduce the computing power requirements of the original spatial-temporal graph convolution network. The masks are pre-trained with the train dataset, which is named "trainable mapping" in Fig.\ref{fig:attProc}. For example in node dimension, the history load input $\chi_t^{r-1}$ will be processed according to: 

\begin{align} 
\label{equ:attention}
S\ =\ V_s\ \times\sigma\left(\chi_t^{r-1}W{\chi_t^{r-1}}^T+b_s\right), 
\end{align} 
where $\chi_t^{r-1}=\left(X_1,X_2,\ldots,X_{T_{r-1}}\right)\in \mathbb{R}^{N\times T_{r-1}}$, $r$ is the layer index, $N$ is node number, $T_{r-1}$ is time period length of the $r-1$ layer, $W\in \mathbb{R}^{N\times T_{r-1}}$, $b_s\in \mathbb{R}^{N\times N}$. $V_s$, $W$, $b_s$ are trainable parameters, and we use $sigmoid$ as an activation function. Attention masks are dynamically computed for different inputs, $S_{i,j}$ represents the correlation strength of node $i$ and node $j$. A normalization is introduced to $S$ rows (Equ.\ref{equ:normalization}).

\begin{align} 
\label{equ:normalization}
{S^{i,j}}^\prime=\frac{exp\left(S^{i,j}\right)}{\sum_{j=1}^{N}exp\left(S^{i,j}\right)}
\end{align}

The internal procedure of the attention block is shown in Fig.\ref{fig:attProc}. The input historical nodal loads are consecutively re-weighted (can be regarded as a 'mapping') first by temporalAtt and then spatialAtt mask based on their co-importance to forecasting LMP. More specifically, the input of the Attention Block will generate two attention masks ($N~nodes \times N~nodes$ for SpatialAtt Mask and $T~hours \times T~hours$ for TemporalAtt Mask), and then the input would do dot-product with the two masks one after another to obtain a mapped input.

\subsection{ST-Conv Block}
To capture both patterns in time and space, a spatial-temporal convolutional (ST-Conv) block structure (ST-Conv block in Fig.\ref{fig_zoomast}) is adopted in this model. It contains two modules: the first is a spectral graph convolution layer to extract the spatial features of LMP, then the second is a traditional convolution in the time dimension to exploit history dependencies for each node. 

\subsubsection{Graph Convolution}
The spectral graph theory generalizes the convolution operation from grid-based data to graph structure data and accelerates it with spectral techniques. The power system is naturally one of such graphs. As for the graph, we applied ChebNet\cite{defferrard2016convolutional} to the graph convolution layer, which uses Chebyshev polynomial to reduce its computational costs and to accelerate the convolution process. The GCN layer based on ChebNet can be represented as follows:
\begin{align}
    y=\sigma\left(g_\theta\ast x\right)=\sigma\left(\sum_{k=0}^{K-1}\theta_{k}T_K\left(\widetilde{L}\right)x\right)
\end{align}
where $x$ denotes the layer input, $y$ denotes the output. $\widetilde{L}$ is the convolution kernel whose parameters rest with training and $\widetilde{L}=\frac{2L}{\lambda_{max}}-I_N=U\widetilde{\Lambda}xU^T$. $K$ represents a $K^{th}$ truncation of Chebyshev polynomial, and this also leads to a K-step receptive field of graph convolution. $T_K$ is a Chebyshev polynomial $T_k\left(x\right)=2xT_{k-1}\left(x\right)-T_{k-2}\left(x\right)$, where $T_0\left(x\right)=1$, $T_1\left(x\right)=x$.
\begin{align} 
    \mathcal{X}_{h}^{(r)}=\operatorname{ReLU}\left(\Phi *\left(\operatorname{ReLU}\left(g_{\theta} *_{G} \hat{\mathcal{X}}_{h}^{(r-1)}\right)\right)\right) \in \mathbb{R}^{ N \times T}
\end{align}
where $\mathcal{X}_{h}^{(r)}$ represents the output of the $r^{th}$ layer, $*$ denotes a standard convolution operation, $\Phi$ is the trainable parameters of the temporal dimension convolution kernel, and the activation function is $\operatorname{ReLU}$. Since the data along the temporal dimension are aligned (a.k.a. Euclidean), a standard convolution is enough to extract the potential influence from previous data.

\subsection{Fully Connected Layer}
The output of the ST-Conv block does not meet the shape that forecasting requires, so the output is altered to an appropriate shape with dot-product. As illustrated in Table. \ref{tab:ASTGCN_params}, the ST-Conv block output of shape $T\times N\times128$ is transposed to $ 1\times N\times 128 \times T$ and then multiplied by a corresponding matrix to formulate the shape of $1\times N\times q $. Finally, the three branches are respectively sum-reduced along with $q$, which is of $N\times 1$. The forecasting LMP of each node is a composition of all three branch outputs.
\begin{align}
    LMP = \mathcal{X}_{\lambda} + \mathcal{X}_{s} \mathcal{X}_{\mu} \in \mathbb{R}^{N}
\end{align}
where $\mathcal{X}_{\lambda}$, $\mathcal{X}_{s}$, and $\mathcal{X}_{\mu}$ denote the output of the three branches, which are $\in \mathbb{R}^{N}$.

\section{Training Method}
\label{trainingmethod}

All neural networks need training, details of how the training is set in the proposed model will be discussed in this section.

\subsection{Loss Function}
Loss function plays an important role in model training. The proposed model contains three branches, so the necessary part is to decide how each branch weighs in the overall loss function. We set varying loss functions for different branches as below:
\begin{align}
    loss_{energy}= \Vert\lambda_{pred}-\lambda_{GT} \Vert_1+ \Vert\lambda_{pred}-\lambda_{GT} \Vert_2 \\
    loss_{congest}= \Vert\mu_{pred}-\mu_{GT} \Vert_1+2 \Vert\mu_{pred}-\mu_{GT} \Vert_2 \\
    loss_{status}=cross\_entropy\left(s_{pred},s_{GT}\right)
\end{align}
where $\Vert\cdot\Vert_1$ is 1-norm, $\Vert\cdot\Vert_2$ is 2-norm. We set a target loss function weighing the ones above for the training process:
\begin{align}
    loss_{total}=loss_{energy}+10\cdot loss_{congest}+100\cdot loss_{status}
\end{align}

\subsection{Parameter Initialization}
The network contains a lot of parameters to be trained, or namely trainable parameters. They need to be pre-set before the training begins, which is called weight initialization. In deep learning networks, it could determine the layer outputs during the course of a forward pass through the network. If either the outputs' vanishing or exploding occurs, loss gradients will either be too large or too small to flow backwards beneficially, and the network will take longer to converge, if it is even able to do so at all.

In the proposed network, Xavier Initialization \cite{glorot2010understanding} is applied, which sets a layer’s weights to values chosen from a random uniform distribution that’s bounded between $\pm \frac{\sqrt{6}}{\sqrt{n_{i}+n_{i+1}}}$, where $n_i$ is the number of incoming network connections, or “fan-in” to the layer, and $n_{i+1}$ is the number of outgoing network connections from that layer, also known as the “fan-out”. According to Glorot and Bengio, Xavier Initialization can maintain the variance of activations and back-propagated gradients up or down the layers of a network and therefore brings substantially faster convergence.

\subsection{Training Settings}
The proposed LMP forecasting model is implemented with TensorFlow 1.14. The training process involves multiple hyperparameters. We tested the number of the terms of Chebyshev polynomial $K \in \{1,3,5\}$, and the accuracy is exalted with $K$ rising. However, the computing cost increased rapidly, so for a better trade-off in both forecasting performance and computing efficiency, we set $K=3$ Similarly, the kernel size of time-convolution is also set to $3 \times 1$. The model is optimized using Adam Optimizer and the initial learning rate is set to 1e-4 for 100 epochs.

This work is examined by comparing MLP, GCN and ASTGCN. The following Fig. \ref{fig:inout} shows how the outputs are generated with the corresponding inputs. The GCN only takes in the latest loads to give a forecast of LMP for the next period, while  ASTGCN (adding temporal Conv and then attention) gives the forecast using previous $T$ hours.

\begin{figure}[htbp]
  \centering
  \includegraphics[width=0.7\linewidth]{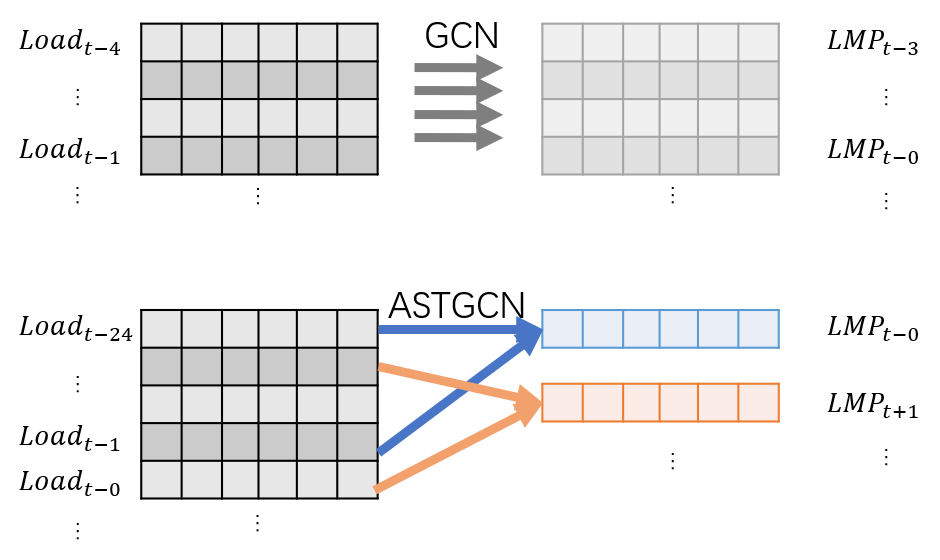}
  \caption{Input and Output of GCN/ASTGCN}
  \label{fig:inout}
\end{figure}

\section{Case Study}
\label{case}
\subsection{Dataset Description}

The dataset involved in this work contains the topology of one IEEE-118 power system. The historical hourly loads and LMPs of all 118 nodes within 3 years are also included. It is divided into a train set and a test set with a proportion of 2:1. To be more specific, the first 2 years are set aside for training and the remaining 1 year is for testing. 

\begin{table}[htbp]
	\centering
	\caption{Dataset Details}
	\label{tab:deeplearning:dataset}
	\begin{tabular}{ccccc}
		\toprule[1.5pt]
		Topology & Vertices & Freq & Time Span \\
		\midrule[1pt]
		IEEE-118  & 118 & 1 point/hr & 3yr\\
		\bottomrule[1.5pt]
	\end{tabular}
\end{table}

The dataset originates from the IEEE 118 case (of which the power line topology is given). We select real load data from the 2016-2018 PJM market~\cite{pjm}, including 26 load areas. The other 92 nodes' loads are generated by linear-weighing the data above and adding noise as following:

For moment $t$, denote the known load data by $d_t^{26\times1}$, then the generated load for the other 92 areas $\widetilde{d}_t^{92\times1}$ are:
\begin{align}
    \label{equ:dataset:loads:dirichlet}
    \widetilde{d}_t^{92\times1}=W^{92\times 26}{d_t^{26\times1}}+\alpha {{N}^{92\times 1}}\left( 0,1 \right)
\end{align}
where $W^{92\times 26}$ is a weight matrix obtained by sampling with Dirichlet distribution, which satisfies: 
\begin{align}
    \label{equ:dataset:loads:sum}
    \sum_{j}W_{i, j}=1, \forall i 
\end{align}

To induce system congestion, we add transmission capacity constraints to the highest mean-transmission-power lines. Also, we assume that bid curves are of quadratic function and are subject to stochastic noises. For generator $i$ at moment $t$ with active power output $g_{i,t}$, the bidding function is:
\begin{align}
    C_i\left(g_{i,t}\right)=c_{2,i}\left(t\right)g_{i,t}^2+c_{1,i}\left(t\right)g_{i,t}
\end{align}
where $c_{2,i}$, $c_{1,i}$ are coefficients depending on time and noises:
\begin{align}
    c_{2,i}\left(t\right)=\frac{1}{1000}\sum_{j}{d_{j,t}c_{20,i}}+0.001\mathcal{N}\left(0,1\right)\\
    c_{1,i}\left(t\right)=\left(0.5+\frac{1}{50000}\sum_{j} d_{j,t}\right)c_{10,i}+0.5\mathcal{N}\left(0,1\right)
\end{align}
where $c_{20,i}$, $c_{10,i}$ are bidding coefficients of generator $i$ in IEEE 118, $\mathcal{N}\left(0,1\right)$ is standard normal distribution.

By solving the economic dispatch problem, we could acquire a dataset containing what is needed to train and evaluate the proposed model, which is separated into two parts: 2016-2017 as train dataset and 2018 as test dataset. In the case study, we suppose the model has acquired all history loads for each node as required in Fig. \ref{fig:inout} and tries to forecast the upcoming LMP at each node in the power system.

With the generated dataset above, we compare the performance of load forecasting among traditional MLP (Multi-Layer Perceptron), GCN, and ASTGCN (GCN with temporal convolution and attention mechanism).

\subsection{MLP v.s. GCN}

\begin{figure*}[htbp]
  \centering
  \includegraphics[width=1\linewidth]{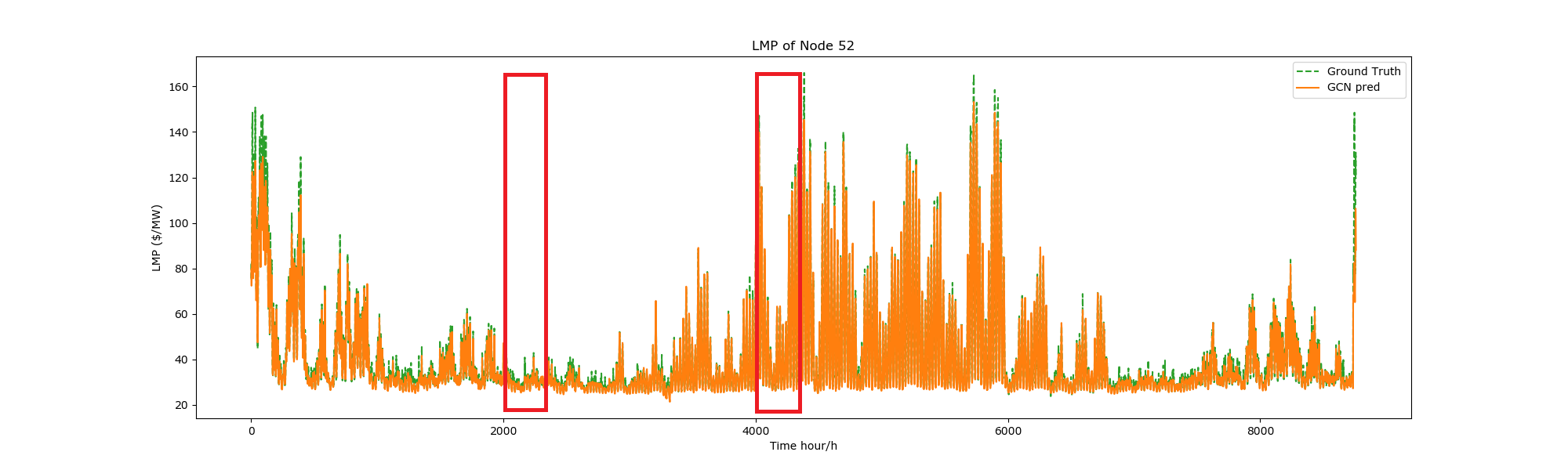}
  \caption{Predictions of GCN and Ground Truth at node 52}
  \label{fig:expriment_result:LMP_52_year}
\end{figure*}

The performance of GCN on the test dataset is shown in Table.\ref{tab:experiment_result:metrics_GCN}. To observe how GCN performs at a single node by time, we take the forecasted LMP curve at node 52 as an example in Fig.\ref{fig:expriment_result:LMP_52_year}.

\begin{table}[h]
  \centering
  \caption{GCN performance on test dataset}
  \label{tab:experiment_result:metrics_GCN}
  \begin{tabular}{*{3}{c}}
    \toprule[1.5pt]
    MAE/(\$/MWh) & RMSE/(\$/MWh) & MAPE/(\%) \\
    \midrule[1pt]
    0.6750 & 1.352 & 1.691 \\
    \bottomrule[1.5pt]
  \end{tabular}
\end{table}

\begin{figure*}[htbp]
  \centering
  \includegraphics[width=1\linewidth]{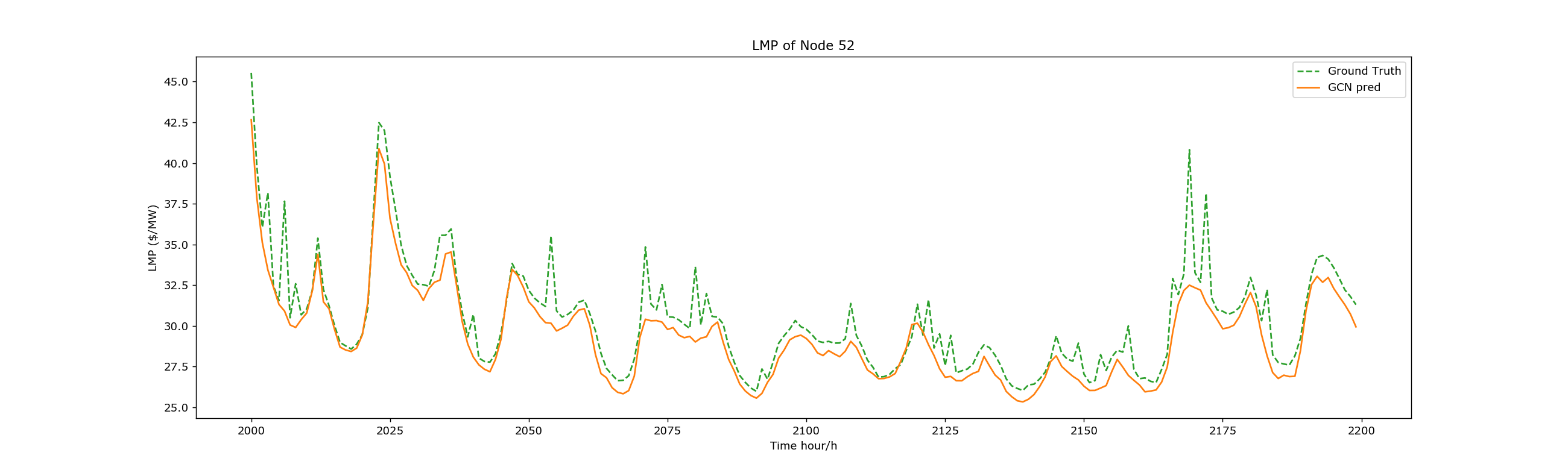}
  \caption{Zoom-in 1 of Ground Truth and GCN's predictions at node 52}
  \label{fig:expriment_result:LMP_52_year_part1}
\end{figure*}

\begin{figure*}[htbp]
  \centering
  \includegraphics[width=1\linewidth]{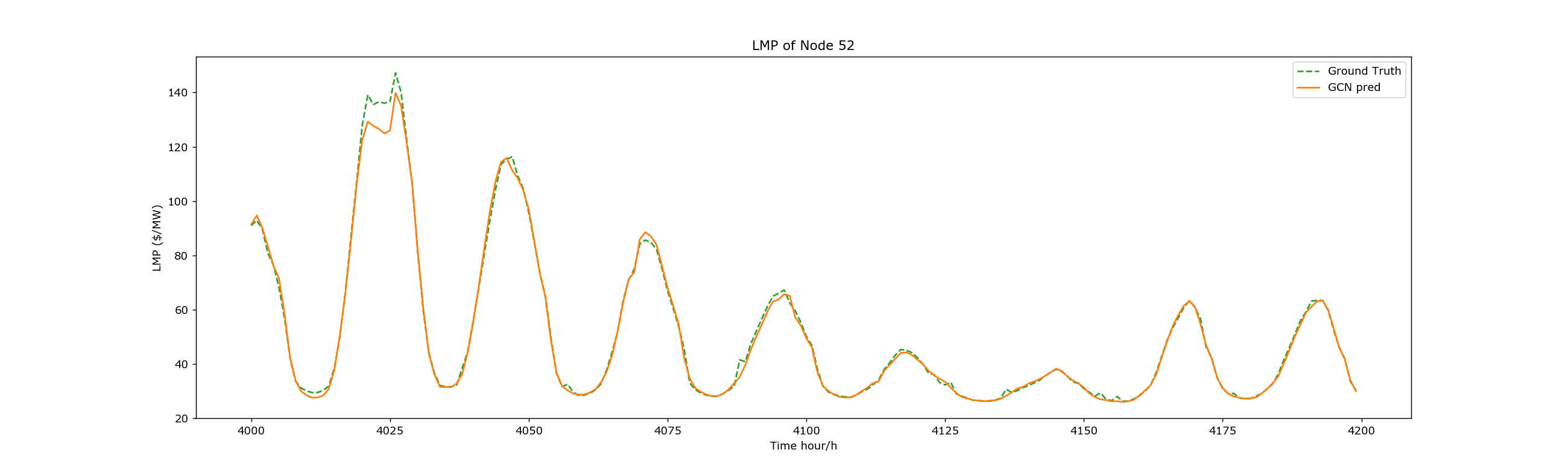}
  \caption{Zoom-in 2 of Ground Truth and GCN's predictions at node 52}
  \label{fig:expriment_result:LMP_52_year_part2}
\end{figure*}

With focused sight into the red boxes on the LMP curve, we get detailed prediction performance visualizations in Fig.\ref{fig:expriment_result:LMP_52_year_part1} and Fig.\ref{fig:expriment_result:LMP_52_year_part2}. It can be induced from the figures that GCN could make precise LMP forecasting, especially when the fluctuations are gentle like that in Fig.\ref{fig:expriment_result:LMP_52_year_part2}.

\begin{figure}[h]
  \centering
  \includegraphics[width=0.8\linewidth]{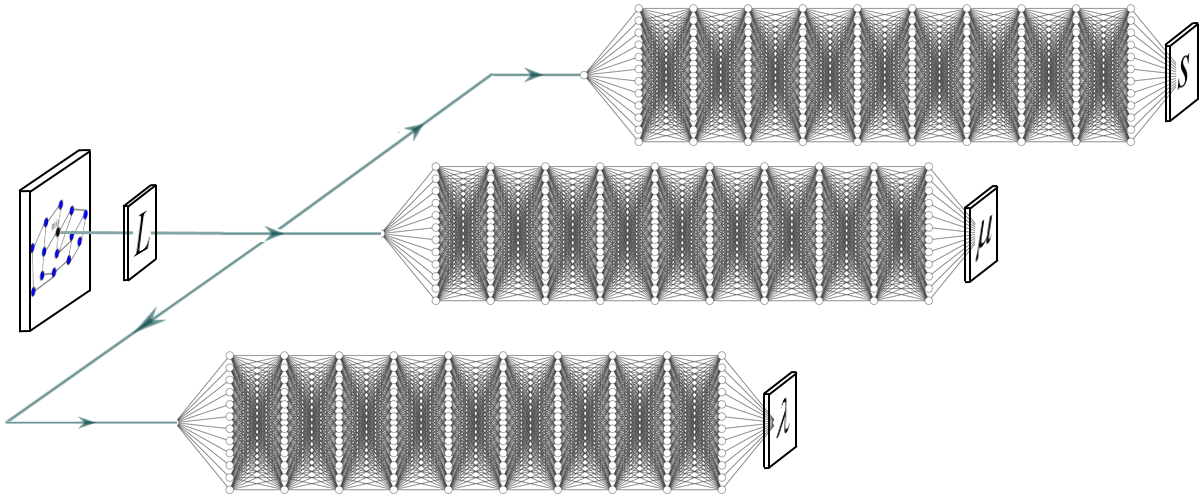}
  \caption{MLP structure}
  \label{fig:MLPstructure}
\end{figure}

We then construct a Multi-Layer Perceptron (MLP) based model to represent what is widely adopted in prevailing LMP forecasting works for comparison. The disadvantage of MLP is that for each node a separate model needs to be constructed to forecast its LMP, and this also hampers MLP to find potential relations between different nodes. To compare fairly, a structure of MLP network resembling that of GCN with three branches is proposed as in Fig.\ref{fig:MLPstructure}. The MLP uses 10 hidden layers with 128 neurons for each. We randomly select node 21, 49, 52 ,76 ,85, and 101 and train one MLP model for each nodes' LMP forecasting. The comparison between MLP and GCN including MAE and RMSE is shown in Table.\ref{tab:experiment_cmp:MAE_GCN_MLP} and Table.\ref{tab:experiment_cmp:RMSE_GCN_MLP}, which strongly demonstrates GCN's effectiveness as it achieves much lower errors in terms of both metrics.

\begin{table}[htbp]
  \centering
  \caption{MAE comparison of GCN and MLP}
  \label{tab:experiment_cmp:MAE_GCN_MLP}
  \begin{tabular}{c*{3}{r}}
    \toprule[1.5pt]
    Node Index & GCN & MLP & Improve(\%) \\
    \midrule[1pt]
    21  & \textbf{1.024} & 4.480 & 81.38 \\
    49  & \textbf{1.251} & 2.706 & 74.48 \\
    52  & \textbf{1.071} & 3.150 & 77.12 \\
    76  & \textbf{1.111} & 1.211 & 55.01 \\
    85  & \textbf{1.016} & 1.176 & 53.34 \\
    101 & \textbf{1.061} & 2.021 & 70.13 \\
    \bottomrule[1.5pt]
  \end{tabular}
\end{table}

\begin{table}[htbp]
  \centering
  \caption{RMSE comparison of GCN and MLP}
  \label{tab:experiment_cmp:RMSE_GCN_MLP}
  \begin{tabular}{c*{3}{r}}
    \toprule[1.5pt]
    Node Index & GCN & MLP & Improve(\%) \\
    \midrule[1pt]
    21  & \textbf{1.427} & 7.910 & 81.88 \\
    49  & \textbf{1.750} & 5.130 & 75.87 \\
    52  & \textbf{1.623} & 6.777 & 79.40 \\
    76  & \textbf{1.597} & 2.004 & 43.16 \\
    85  & \textbf{1.508} & 1.934 & 43.64 \\
    101 & \textbf{1.570} & 3.874 & 69.72 \\
    \bottomrule[1.5pt]
  \end{tabular}
\end{table}

According to the experiment, GCN shows a prominent advantage over the traditional MLP methods in LMP forecasting, including its great improvement in precision and simplification (as we need only one model for all nodes with GCN).

\subsection{GCN v.s. ASTGCN}
GCN could extract the topological correlations among nodes, but the history data of nodes' loads are utterly omitted. However, in reality, previous loads’ trends usually have some impact on future LMP, on which basis traditional statistical models were built. Thus, an enhanced GCN with Temporal Convolution and Attention Mechanism taking historical data into account is proposed in this work.

\begin{figure}[htb]
  \centering
  \includegraphics[width=0.7\linewidth]{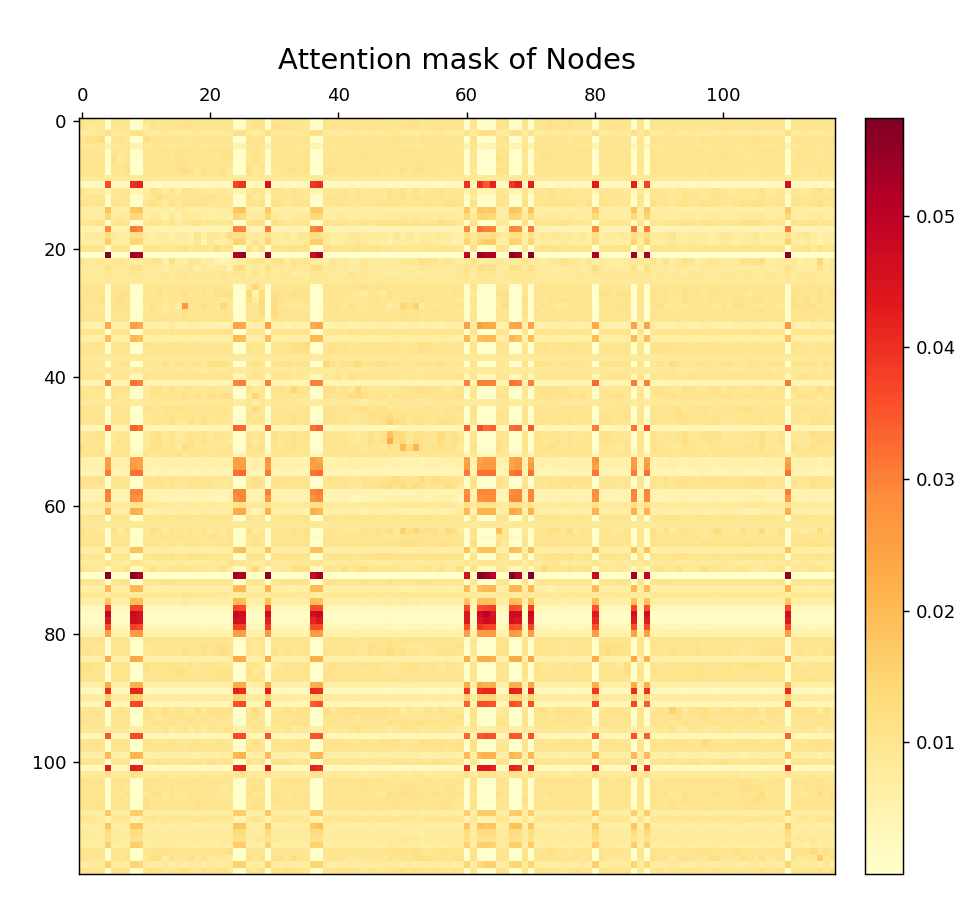}
  \caption{Attention Mask of Nodes}
  \label{fig:attnode}
\end{figure}

\begin{figure}[htb]
  \centering
  \includegraphics[width=0.7\linewidth]{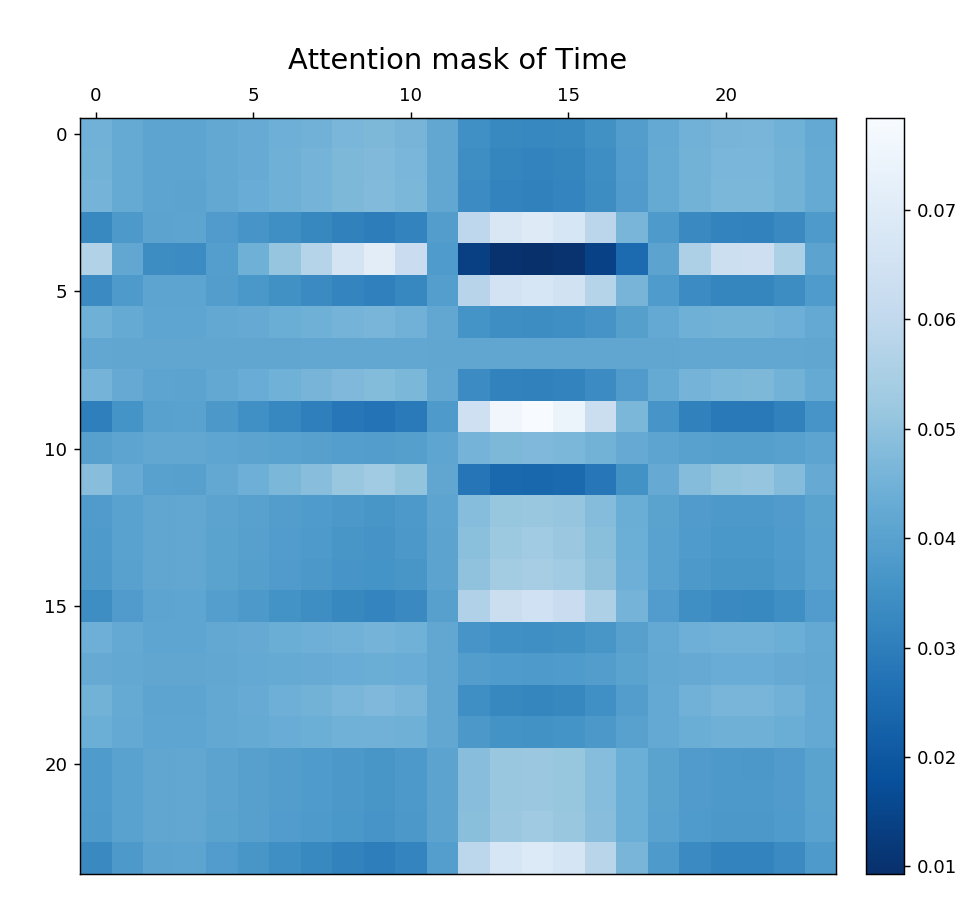}
  \caption{Attention Mask of Time}
  \label{fig:atttime}
\end{figure}

\begin{figure}[htb]
  \centering
  \includegraphics[width=0.8\linewidth]{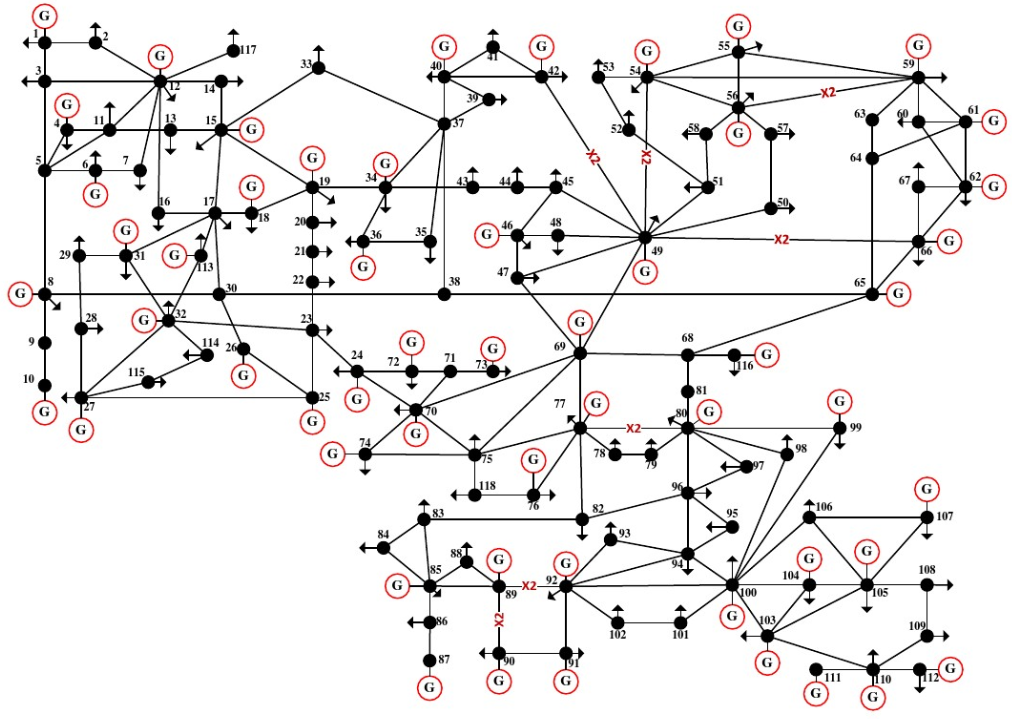}
  \caption{IEEE118 Topology}
  \label{fig:ieee118}
\end{figure}

The introduced Graph Convolution takes effect in extracting topological information, while the added temporal convolution tries to discover the influence of time-continuity of power loads. In addition, the purpose of borrowing attention mechanism is to weigh historical node loads differently, meanwhile exalting the inter-weights of highly related nodes. Fig.\ref{fig:attnode} shows a group of typical attention masks (illustrated in Section \ref{attmech}) among nodes generated with one input. The colored square of row $i$ column $j$ represents the influence of $j$th node on $i$th node (the redder means the higher importance). Thus, along the column axis, it is easy to discover some nodes are of higher influence on other nodes (like 5, 9, 10, 25, 26, 30, 37, 38, 61, 63, 64, 65, 68, 69, 71, 81, 87, 89, 111). From the topology of IEEE 118 (Fig. \ref{fig:ieee118}), most of these nodes are located in crossroads of power lines or at the only neighbour of generators. Some of them are not that special, while their high attention scores indicate their significance in LMP prediction. This feature shows an interpretable advantage of the proposed model highlighted hot nodes and connections' strength among them. Fig.\ref{fig:atttime} shows the attention scores among different periods in the same way.

\begin{figure}[htb]
  \centering
  \includegraphics[width=1\linewidth]{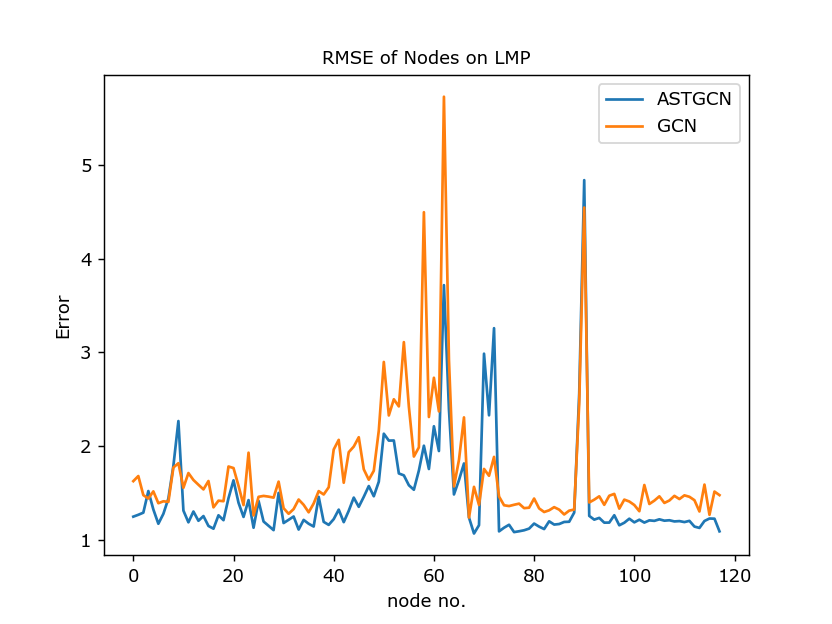}
  \caption{RMSE of Nodes on LMP by GCN and ASTGCN}
  \label{fig:rmse}
\end{figure}

\begin{figure*}[htbp]
  \centering
  \includegraphics[width=1\linewidth]{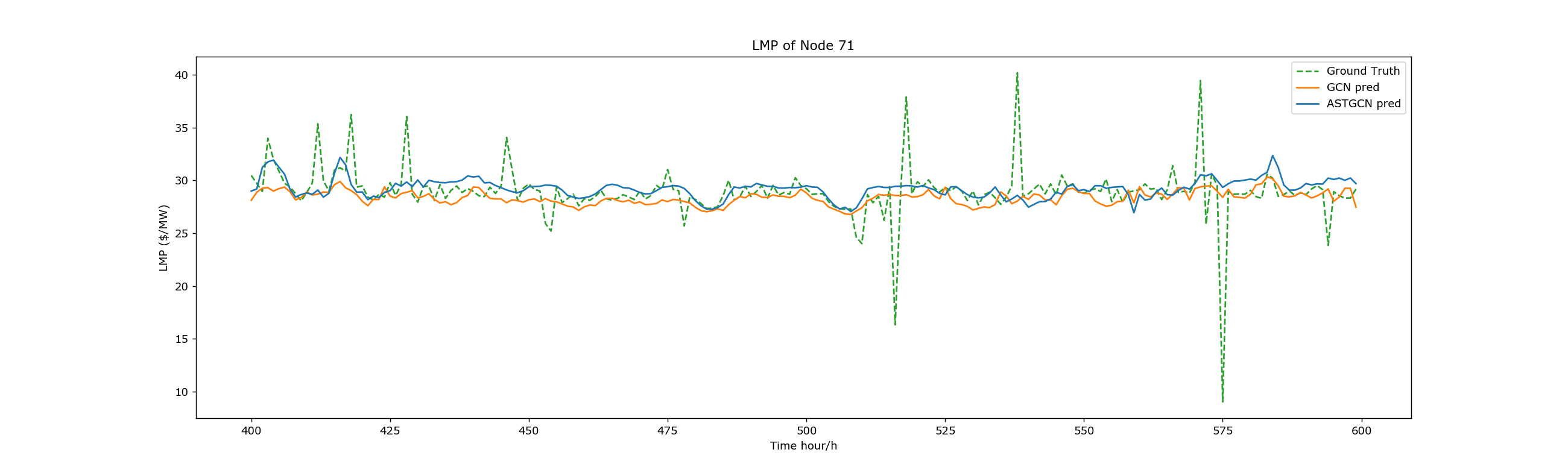}
  \caption{Bad case of ASTGCN}
  \label{fig:badcase}
\end{figure*}

ASTGCN brings the model reduced RMSE at most nodes (shown in Fig. \ref{fig:rmse}), showing a consistent superiority of ASTGCN in most cases. At nodes 70, 71 and 72, however, the ASTGCN fails to give a better prediction. This might comes from the accumulating errors when there occurs a highly fluctuating LMP curve. Since the ASTGCN additionally (compared with GCN) considers the historical loads, its tries to formulate a smooth temporal trend and this goes in the wrong direction for such cases and induces a bad RMSE at such nodes like node 71 (Fig. \ref{fig:badcase}) with strong temporal fluctuations.

\begin{figure*}[htbp]
  \centering
  \includegraphics[width=1\linewidth]{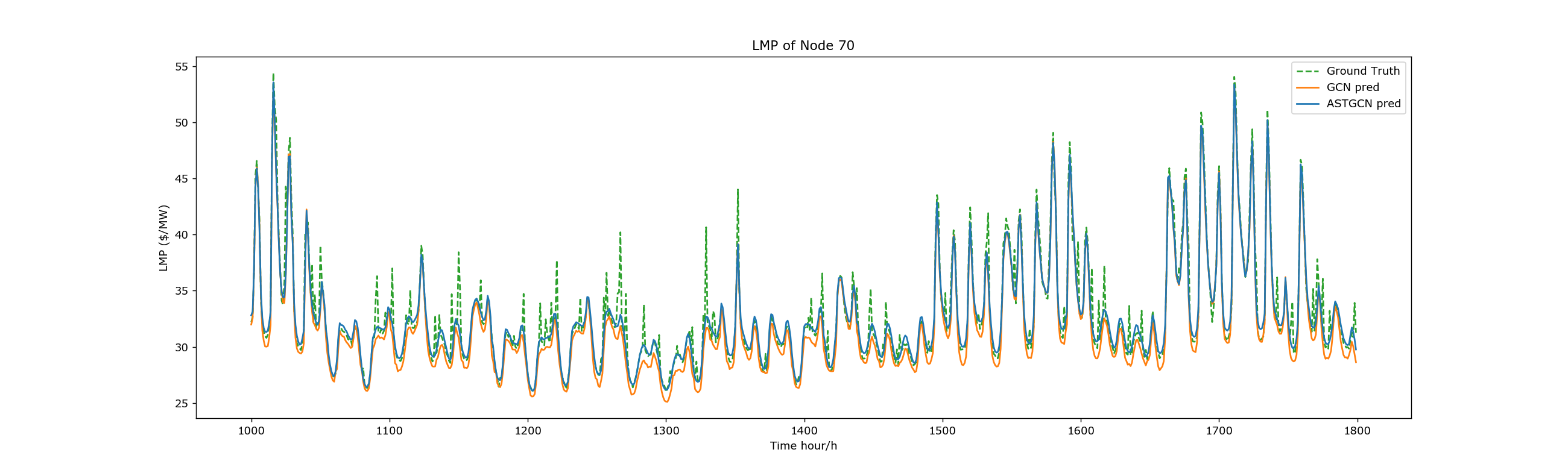}
  \caption{Performance Contrast 1 of GCN and ASTGCN}
  \label{fig:astgcn1}
\end{figure*}

\begin{figure*}[htbp]
  \centering
  \includegraphics[width=1\linewidth]{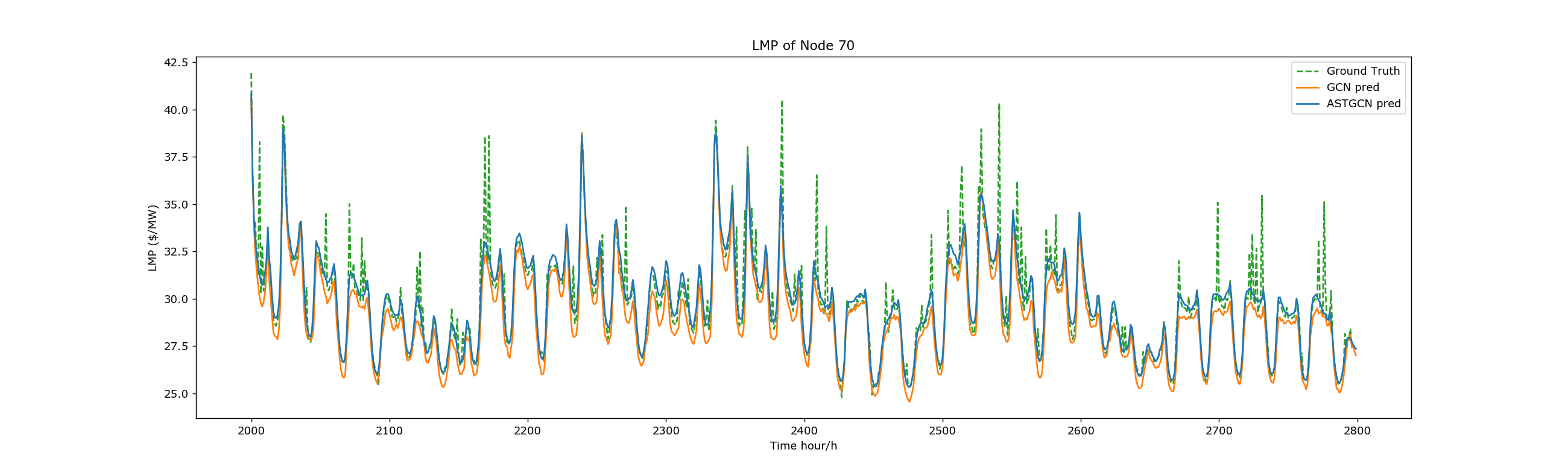}
  \caption{Performance Contrast of GCN and ASTGCN}
  \label{fig:astgcn2}
\end{figure*}

A comparison of performance is shown in Table.\ref{tab:experiment:compare}. We can see generally a progressive advance in both MAE and RMSE is achieved via ASTGCN, while the accuracy of congestion factor $s$ remains almost the same.
\begin{table}[htbp]
  \centering
  \caption{Comparison of GCN and ASTGCN}
  \label{tab:experiment:compare}
  \setlength{\tabcolsep}{2mm}{
  \begin{tabular}{cccc}
    \toprule[1.5pt]
    Model & \makecell[c]{Accuracy of\\congestion factor $s$} & \makecell[c]{MAE} & \makecell[c]{RMSE} \\
    \midrule[1pt]
    baseline GCN & 93.8242\% & 0.987564 & 1.926941 \\
    ASTGCN  & 93.6758\% & \textbf{0.822848} & \textbf{1.538259}\\
    \bottomrule[1.5pt]
  \end{tabular}}
\end{table}

In a nutshell, the attention could adaptively re-weigh the load according to their importance, and as well provide some interpretability of how we get some LMP prediction. Meanwhile, the temporal convolution puts enough emphasis on the previous load information which benefits the model by fusing traditional time series models.

\section{Conclusion}
\label{conclusion}
To utilize both the system topology and time series of power loads in LMP forecasting, this paper proposes a novel LMP forecasting method based on the GCN. Several improvements are promoted, including the spatial-temporal convolution and attention mechanism. A three-branch network is introduced to predict the respective components of LMP. The case study shows that the proposed GCN-only model outperforms the existing MLP in both accuracy and simplicity by an average of 30\% - 40\% in prediction errors. With the ST-Conv blocks and attention blocks, ASTGCN succeeds to capture the dynamic spatial-temporal characteristics of LMPs. Further experiments on IEEE 118 dataset shows its capability to utilize more information and enhance precision.

Note that the GCN-based LMP forecating method can also be extended to similar applications that involve other time series related to system topological structure. Still, the proposed method reaches its bottleneck when it comes to frequent LMP spikes. This might comes from the inherence of convolution operations, and more efforts in network designing are expected to tackle such defects.

\section{Acknowledgements}
This work was supported by the National Key R\&D Program of China under Grant No. 2020YFB0905900.

\bibliographystyle{iet}
\bibliography{ref}

\end{document}